\documentclass{article}
\usepackage{ijcai22}

\usepackage{times}
\usepackage{soul}
\usepackage{url}
\usepackage[hidelinks]{hyperref}
\usepackage[utf8]{inputenc}
\usepackage[small]{caption}
\usepackage{graphicx}
\usepackage{amsmath}
\usepackage{amsthm}
\usepackage{booktabs}
\usepackage{algorithm}
\usepackage{algorithmic}
\usepackage{color}
\usepackage{makecell}
\usepackage{indentfirst}
\usepackage{stfloats}
\newcommand{\ignore}[1]{{}}

\newcommand{\pre}{\textit{pre}}
\newcommand{\eff}{\textit{eff}}

\newcommand{\ER}{\textsf{R}_e}
\newcommand{\RR}{\textsf{R}_r}

\newcommand{\RI}{\textsf{R}_i}

\newcommand{\ARMS}{\texttt{ARMS}}
\newcommand{\AMAN}{\texttt{AMAN}}
\newcommand{\AMDN}{\texttt{ARDN}}
\newcommand{\FAMA}{\textsf{FAMA}}

\newcommand{\our}{\texttt{TAMP}}
\newcommand{\SPM}{\texttt{DSPM}}
\newcommand{\TMM}{\texttt{TMM}}
\newcommand{\ablation}{\texttt{TAMP}$^-$}

\newcommand{\Extractor}{\delta}
\newcommand{\Rules}{\textsf{ApplyRule}}
\newcommand{\ActionLearner}{\textsf{A}}

\newcommand{\Vocabulary}{\textsf{V}}
\newcommand{\Topic}{\textsf{T}}
\newcommand{\Encoder}{\mathbf{E}}
\newcommand{\Decoder}{\mathbf{D}}
\newcommand{\Repara}{\mathbf{R}}

\title{Text-Based Action-Model Acquisition for Planning}

\author{
Kebing Jin \footnotemark[2]
\and
Huaixun Chen \footnotemark[2] \and
Hankz Hankui Zhuo \footnotemark[1]
\affiliations
School of Computer Science and Engineering, Sun Yat-sen University, Guangzhou, China\\
\emails
jinkb@mail2.sysu.edu.cn,
chenhx59@mail2.sysu.edu.cn,
zhuohank@mail.sysu.edu.cn
}

\begin{document}

\maketitle
\renewcommand{\thefootnote}{\fnsymbol{footnote}}
\footnotetext[1]{Corresponding author}
\footnotetext[2]{These authors contributed equally}
\begin{abstract}
    Although there have been approaches that are capable of learning action models from plan traces, there is no work on learning action models from textual observations, which is pervasive and much easier to collect from real-world applications compared to plan traces.
    In this paper we propose a novel approach to learning action models from natural language texts by integrating Constraint Satisfaction and Natural Language Processing techniques. Specifically, we first build a novel language model to extract plan traces from texts, and then build a set of constraints to generate action models based on the extracted plan traces. After that, we iteratively improve the language model and constraints until we achieve the convergent language model and action models. We empirically exhibit that our approach is both effective and efficient. 
\end{abstract}
\vspace{-5mm}
\section{Introduction}
Automated Planning aims at synthesizing plans to transit initial states to goals. 
Generating plans requires specification of domains in the form of planning language, such as PDDL \cite{PDDL}.
It is often tedious or difficult to build domains by hand due to the high requirement of manual efforts and domain knowledge. Automatically learning domain models has significantly attracted researchers' attentions recently \cite{ARMS,LOCM,DBLP:journals/ai/ZhuoK17,DBLP:conf/ijcai/LamannaSSGT21}.  


Despite the success of previous domain model learning approaches, they are restricted to learn from correct plan traces. 
To overcome the limitations, some methods were proposed to capture the relations between observed noisy actions or noisy states, such as {\AMAN} \cite{AMAN}, and {\AMDN} \cite{AMDN}. 
In practice, however, it is difficult to construct states which are made up of propositions for human without expert knowledge, let along creating relations from unstructured data, such as images and texts, applying to all areas of human life.
Compared with structured data in the form of state or action sequences, unstructured data, e.g., natural language descriptions, is variable and often along with omitting, resulting in difficulties in reasoning about the observations and picking up the implied rules. 
In order to capture the regularities of changing from unstructured data, there have been works on acquiring action models from natural language descriptions, such as text instructions and summaries \cite{DBLP:conf/icaart/YordanovaK16,DBLP:conf/aips/LindsayRFHPG17,storyframer,DBLP:conf/aaai/HaytonPFL20}. 
Most of them first extract operators and objects by deriving linguistic annotations, analyses the relations of operators and generate PDDL models. 
These methods extract actions, even states, from natural language description of actions. 
Despite they can speculate about causal relationship from the order of action occurring, it is hard to infer complete preconditions and effects of actions without description of observations. 


Learning models via deep learning methods is in great demand recently. It is, however, often hard to train neural planning models for generating valid solutions to planning problems. 
Compared with neural models, human does not only use existed information, but also summarize them and create new knowledge, such as common rules formalized by action models, for helping solving future tasks. 
However, when facing unstructured data like texts, it is challenging to recognize and abstract different sentences with the same meanings without labels. One way to make them be understandable for agents is to formalize them structurally, such as by propositions. The predicate names of propositions, however, may not be involved in the sentences. On the other hand, noisy state traces make action models learning become rather difficult. Meanwhile, lacking accurate action models as guidance results in difficulties correcting proposition mapping. 

\begin{figure}[!ht]
    \centering
    \includegraphics[width=0.48\textwidth]{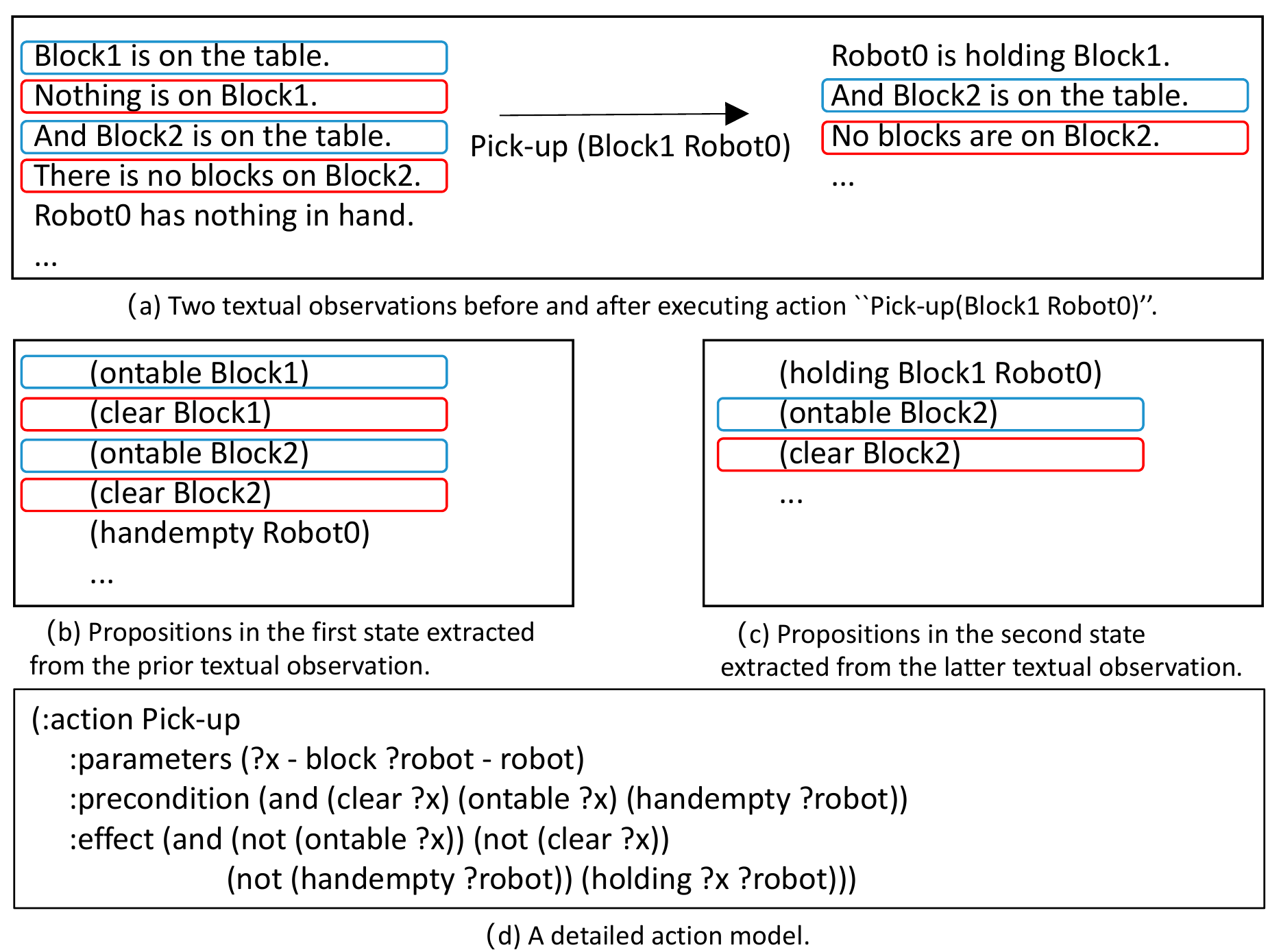}
    \vspace{-3mm}
    \caption{An example of block domain.}
    \label{fig:example}
    \vspace{-5mm}
\end{figure}
\emph{For example, two textual observations before and after executing action ``Pick-up (Block1 Robot0)'' are shown in Figure \ref{fig:example}(a). Propositions of two texts are enumerated in Figure \ref{fig:example}(b) and (c), respectively. We use red blocks to highlight the sentences describing ``(clear ?x - Block)'' and they are written in different ways, where ``clear'' is not included in these sentences. When learning action models, it is difficult to classify these sentences into one group. Besides, distinguishing them from the similar sentences but different meaning is even harder, such as the sentences circled by blue blocks. All these sentences involve ``on'', making they hard to be distinguished. If propositions are extracted inaccurately, learning a correct action model, as shown in Figure \ref{fig:example}(d), is incredibly hard. 
}

To overcome these difficulties, we propose a novel approach, {\our}, standing for \textbf{T}ext-based \textbf{A}ction-\textbf{M}odel acquisition for \textbf{P}lanning. It first maps textual observations to propositional state traces, and constructs action models following PDDL semantics. In particular, we build a variational auto-encoder (VAE) \cite{vae} combined with Embedded Topic Modelling \cite{ETM} to classify sentences into topics and select words as predicate names. Besides, we construct rules to order parameters and form propositions. After that, {\our} improves a neural extractor to reduce the contradiction between action models and traces repeatedly, until converging towards an optimal model. 


\ignore{
We summarize the contributions of the paper as follows:
\begin{itemize}
    \item We build an initializer to map sentences to propositions, even when predicate are not involved in sentences.
    \item We propose a learner to deal with noisy traces, achieved by a neural model inspired by matrix decomposition.
    \item Combining these two methods, we construct an EM-style framework to compute available action models as well as a neural proposition extractor.
\end{itemize}

In the remainder of the paper, we first introduce related works and a problem definition. Then we present {\our} in detail and evaluate {\our} by comparing with baselines in three domains. Finally, we conclude our paper with future works.
}
\vspace{-3mm}
\section{Related Work}
\noindent \textbf{Planning in NLP}
Recently, there have been significant advances in story generation. Text planning is one of crucial steps to guide models to generate well-organized long text. In general, prior studies with text planning usually first planning a sketch and then generate the whole story from the sketch. The sketch can be a sequence of keywords \cite{DBLP:conf/aaai/YaoPWK0Y19,kong-etal-2021-stylized,DBLP:conf/aaai/YuLCY021}, key phrases \cite{DBLP:conf/emnlp/XuRZZC018} or contents \cite{hua-etal-2021-dyploc}. These approaches were able to generate long texts with valid storylines. However, they cannot handle goal-driven tasks, because they cannot reason about the causal relationships between sentences.


\noindent \textbf{Learning action models from plan traces}
Learning action models from plan traces has had a long history. 
Previous methods have made efforts in learning action models with full or partial intermediate states. 
For example, {\ARMS} \cite{ARMS} is able to create STRIPS action models. It defines a set of weighted constraints and uses these constraints to build and solves a weighted propositional satisfiability problem with a MAX-SAT solver. 
{\FAMA} \cite{FAMA} was proposed to learn action models when the actions of the plan executions are partially or totally unobservable and information on intermediate states is partially provided. 
However, these approaches assume that the observed states are observed correctly. 
Considering capturing correct relations between plan traces containing wrong states, Mourao et al. \cite{DBLP:conf/uai/MouraoZPS12} proposed an approach to learn action models with noisy observations of the intermediate state and correct actions. 
{\AMAN} \cite{AMAN}, a graphical model, was proposed to learn action models from noisy plan traces, including actions and states. 
They, however, require the actions to be ordered correctly. 
To achieve that, {\AMDN} \cite{AMDN} captures information from disordered actions, parallel actions, and noisy constraints and make use of a weighted MAX-SAT solver.

\noindent \textbf{Learning action models from texts}
In recent years, researchers have been exploring learning action models from raw texts and numbers of previous works are based on instructional texts. Yordanova et al. \cite{DBLP:conf/icaart/YordanovaK16} extracted verbs and objects from text instructions based on part of speech (POS) tagging module, and discover causal relations on the basis of the order of appearance to build PDDL models. Lindsay et al. \cite{DBLP:conf/aips/LindsayRFHPG17} generated sequences of actions by construct representations of sentences and cluster operators by compute similarity, and build PDDL domain model with the help of a domain model acquisition tool. These works focus on classifying events in instructions and summarizing the causality relationship between them. Some approaches are based on textual descriptions. Sil et al. \cite{DBLP:conf/ranlp/SilY11} used text mining via a search method to identify documents that contain words that represent target verbs or events and uses inductive learning techniques to identify appropriate action preconditions and effects. 
\vspace{-3mm}
\section{Problem Definition}
\vspace{-1mm}
In this paper we aim to capture information from textual observations, formalize them by state traces, and construct knowledge as action models. We first define a textual description by $t = \langle f^0, f^1, \dots, f^j\rangle$ where $f^j$ indicates the $j$th sentence in text $t$. We assume a sentence $f^j$ can be indicated by a proposition $p^j = \langle \rho^j, \Theta^j \rangle$, where $\rho^j$ is a predicate and $\Theta^j$ is a set of parameters, which can be empty. 
Hence, a textual observation $t$ can be abstracted by a state $s = \langle p^0, p^1,\dots, p^j\rangle$. 
It is noted that, in this paper, all parameters can be found in the sentence without omission and extracted by \cite{DBLP:conf/ijcai/FengZK18}. We use $\tau(\theta)$ to denote the type of parameter $\theta \in \Theta$ to indicate its property. As for a proposition $p = \langle \rho ,\Theta \rangle$, we define its topical proposition by $\psi = \langle \rho, \xi\rangle$, where $\xi = \langle \tau(\theta_0),\dots,\tau(\theta_n) \rangle$.

Our training data $\Phi$ is a set of tuples, each one is defined by $\phi = \langle t_0, t_g, \mathcal{T},\sigma \rangle$, where 
$t_0$ is a text describing the initial observations, and $t_g$ stands for goals of the task. $\mathcal{T} = \langle t_0, \dots, t_n\rangle$ is pieces of texts describing observations when handling the task, which requires agents to reach $t_g$ from $t_0$. The neighbouring two texts $t_i$ and $t_{i+1}$ describe a current and following observation before and after executing an action $a_i \in \sigma$ respectively. $\sigma$ = $\langle a_0, \dots, a_{n-1} \rangle $ is a plan in the form of an action sequence, guiding agents to perform the task. Each action is a grounding action model with parameters. \emph{For example, $t_0$ and $t_g$ in the blocks domain are ``Block1 is on table. And Block2 is placed on table too.'' and ``Block2 is on table. And Block2 is under Block1.'' respectively. We can indicate $t_0$ by two propositions, i.e., ``(on-table Block1)'' and ``(on-table Block2)''. Proposition extracted from $t_g$ are ``(on-table Block2)'' and ``(on Block1 Block2)''. The types of ``Block1'' and ``Block2'' are both ``Block''. ``(on-table Block1)'' and ``(on-table Block2)'' relate to a topical proposition ``(on-table ?x - Block)''. $t_0$ and $t_g$ indicate a task asking an agent to pick up a block and place it on another one. To handle the task, $\sigma$ has two actions, i.e., ``Pick-up (Block1)'' and ``Put-down (Block1 Block2)''.}
    
A planning domain $\mathcal{D}$ is composed of action models, each action model is defined by a tuple of $A = \langle a, \pre(a),$ $\eff(a)\rangle$, where $a$ is an action name with zero or more types of parameters. 
$\pre(a)$ is a set of preconditions requiring to be satisfied when executing $a$, each of which is a topical proposition. Similarly, $\eff(a)$ is a set of effects added into or deleted from the state after executing $a$. \emph{For example, ``Pick-up (?x - Block)'' indicates an agent picks up a block, where ``Pick-up'' is an action name and ``?x'' is a parameter whose type is ``Block''. $\pre(a)$ of the action is ``(on-table ?x)'' and ``(hand-empty)'', indicating the agent can pick up a block if it is on the table and the agent holds nothing. $\eff(a)$ is ``(not (on-table ?x))'', ``(not (hand-empty))'' and ``(holding ?x)'', which indicates the block is not on the table but held by the agent.} 

In this paper, we only know names of the executed action models $\mathcal{D} = \langle A_0, A_1, \dots, A_x \rangle$ and the types of parameters they use, which can be found in action sequences $\sigma$. Our aim is to complete the preconditions and effects of action models by capturing information from textual observations.
\vspace{-2.5mm}

\section{Our {\our} Approach}
\begin{figure}[!ht]
    \centering
    \includegraphics[width=0.47\textwidth]{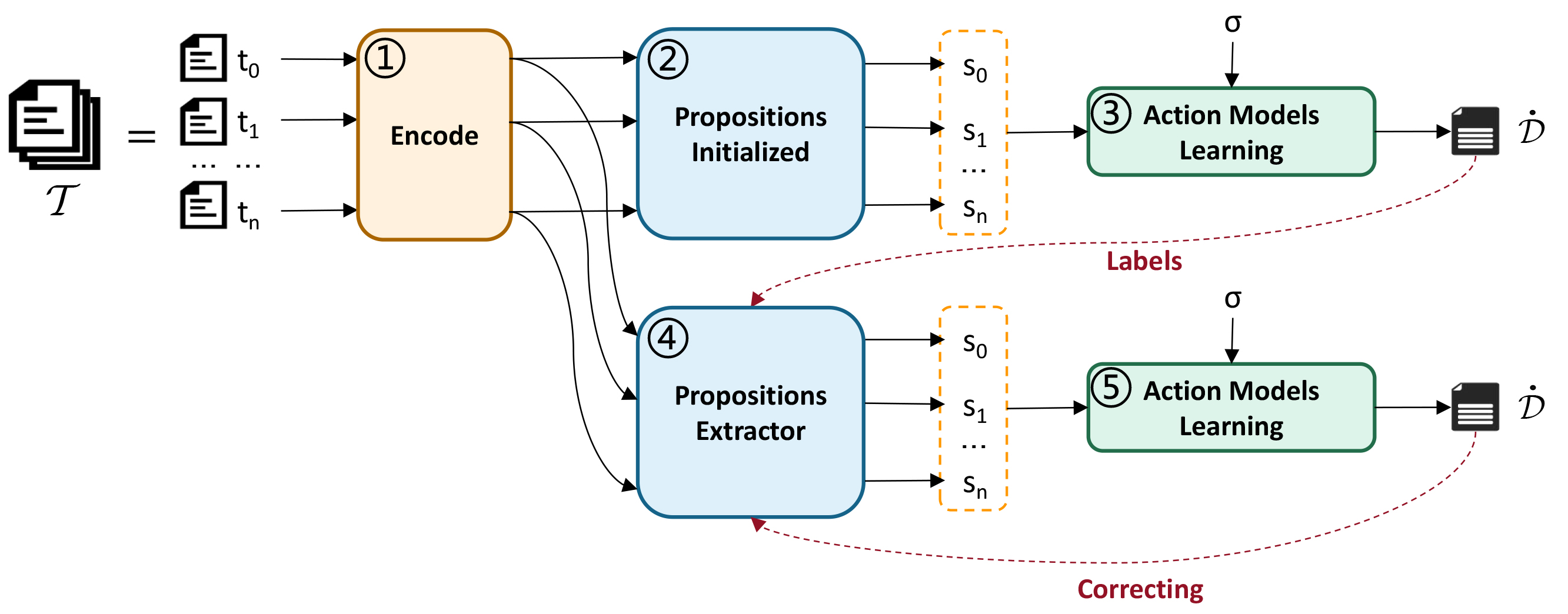}
    \vspace{-1mm}
    \caption{An overview of the proposed {\our}}
    \label{fig:overview}
    \vspace{-5mm}
\end{figure}

In this section, we address {\our} in detail, which includes:
(1) a text encoder to encode textual observation, 
(2) a trace initializer to initialize state traces based on the textual representations, 
(3) an action learner for learning action models from noisy state traces, 
(4) a proposition extractor to transform textual representations to propositional states.
We follow the EM-style (Expectation Maximization) framework, which has been demonstrated effective in learning action models \cite{AMAN}, for alternately optimizing proposition extractor and action learner. 
An overview of {\our} is shown in Figure \ref{fig:overview}. 
We first learn a VAE \cite{vae} to encode sentences (Block \textcircled{1}), then initialize state traces and use them to learn action models (Blocks \textcircled{2} and \textcircled{3}). Next, we learn a mapping from representations of textual observations to propositions, and repeat learning action models and improving the proposition extractor to deal with contradictions between extracted state traces and learned action models (repeating Blocks \textcircled{4} and \textcircled{5}). 

\vspace{-3.5mm}

\vspace{-1mm}
\subsection{Representation Learning and Trace Initializer}
To capture rules involved in textual observations, we first extract information from textual observations, and indicate them by propositional state traces. 
However, we have two major challenges during extracting:
(1) The number of propositions is unknown before learning.
(2) The names of propositions are consistent with human understanding, which may not be from the input documents. How to generate those names satisfying the above-mentioned constraint is a challenging task.
To tackle these challenges, we build a VAE combined with Embedded Topic Modelling \cite{ETM} to encode sentences,
and then use a reparameterization trick to map each sentence to a proper topic. In order to create well-grounded guidance at the beginning of training proposition extractor, we initialize state traces by selecting a word as the name with the highest probability in the topic, i.e., predicate. Then we follow series of rules to organize 
parameters and form them with predicates to be propositions. 


\begin{figure*}[!ht]
    \centering
    \includegraphics[width=0.7\textwidth]{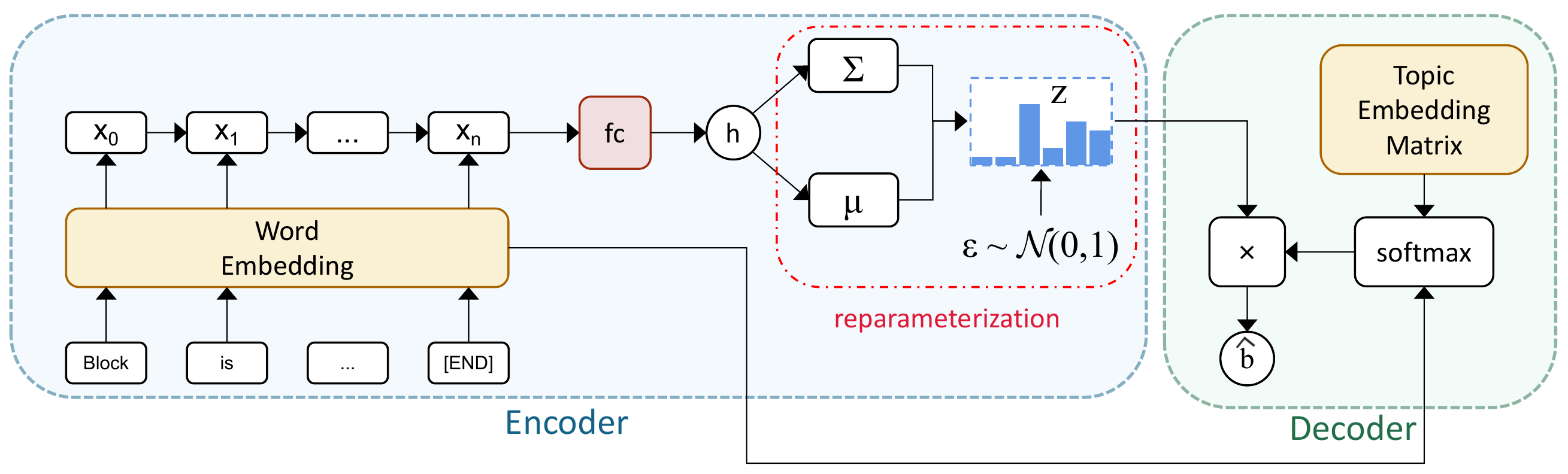}
    \vspace{-1mm}
    \caption{The whole framework of VAE for representation learning, which is composed of an encoder and a decoder.}
    
    \label{fig:propositions_extracting}
    \vspace{-4.5mm}
\end{figure*}

\noindent \textbf{Representation Learning}
We first aim to learn a bidirectional mapping from sentences to hidden states, as shown in the Figure \ref{fig:propositions_extracting}. Considering to encode sentences, a sentence $f$ is first passed through a word embedding neural network to compute embedding for each word. It embeds the vocabulary {\Vocabulary} in a $\iota$-dimension space.
A GRU with a fully connected layer takes the embedding of each word in $f$ as inputs and outputs a hidden state $h$
, denoted by $h = \Encoder(f)$. $h$ is a $\kappa$-dimensional vector, where $\kappa$ is a hyperparameter. After that we use a reparameterization trick to transform a Gaussian random variable to approach a normal distribution, indicating the probability distribution of topics for initializer, denoted by $z = \Repara(h)$.

Decoder further maps $z$ to a bag-of-word representation. 
We first initialize a topic embedding neural network which embeds a topic vocabulary {\Topic} in a $\kappa$-dimensional space.
Topical word embedding is computed by a softmax layer, which takes the product of the word embedding matrix {\Vocabulary} and the topic embedding matrix {\Topic} as an input. It indicates a distributed representation of the words in the space of topics. Finally, the reconstructed bag-of-word representation $\hat{b}$ is computed by $\hat{b} = \Decoder(z) = \textit{softmax}(\Vocabulary \Topic^\top)z$.

The objective of training encoder and decoder is to make reconstructed bag-of-words representation $\hat{b}$ and the original bag-of-words $b$ become as identical as possible. Our loss is defined by Equation (\ref{equation:loss}), composed of two different losses:
\vspace{-1.5mm}
\begin{align}
\label{equation:loss}
L = L_{recon} + L_{KL}
\end{align}
\vspace{-5.5mm}

where $L_{recon}$ is a reconstruction loss $|b - \hat{b}|$ in terms of cross-entropy, and $L_{KL}$ is to minimize the variational loss to ensure our representation to approach a normal distribution, measured by KL divergence loss.
    

\noindent \textbf{Trace Initializer} 
Next, we follow several rules to initialize state traces. 
The reason for building initializer is due to lacking labels for training neural extractor directly.
Despite our framework can improve an untrained extractor, action models learned from noisy traces are probably not of high-quality
. Trace initializer is able to significantly improve the efficiency and effectiveness. We first map sentences to their topics by $z = \Repara(\Encoder(f))$. The likelihood $P(w)$ of each word $w$ to be the predicate of a topic is computed by Equation (\ref{equation:candidate_predicate}), where $w$ with the highest probability is the candidate predicate.
\vspace{-2mm}
\begin{align}
    \notag
    P(w) &= softmax(\Vocabulary \Topic^\top z) \\
    & = softmax(\Vocabulary \Topic^\top \Repara(\Encoder(f)))
    \label{equation:candidate_predicate}
\end{align}
\vspace{-5mm}

To make up a proposition with $w$ and $\Theta$, where $\Theta$ is objects involved in sentence, extracted by \cite{DBLP:conf/ijcai/FengZK18}. We regard them as parameters in propositions. We first initialize an empty topical propositions set $\Psi$ for recording. Given $\Psi$, $P(w)$, and $\Theta$, we construct a proposition following a series of \underline{\textbf{\textsc{Rules}}}, denoted by Equation (\ref{equation:rules}):
\vspace{-0.5mm}
\begin{align}
    p,\Psi = \Rules(\Psi,P(w),\Theta)
    \label{equation:rules}
\end{align}
\vspace{-5mm}

\noindent \underline{\textbf{\textsc{Rule1}}}: If $w$ isn't involved in any propositions in $\Psi$, we define that $p = \langle w, \Theta \rangle$, and $\psi = \langle w, \xi  \rangle$ is its topical proposition where $\xi = \langle \tau(\theta_0),\dots,\tau(\theta_n) \rangle$. Then we add $\psi$ into $\Psi$. 

\noindent \underline{\textbf{\textsc{Rule2}}}: If $w$ is involved in a proposition $\psi = \langle w, \xi\rangle$ where $\psi \in \Psi$, $\xi = \langle \tau(\theta'_0),\dots,\tau(\theta'_n)\rangle$, and there is a one-to-one correspondence letting $ \tau(\theta')  = \tau(\theta)$, 
we define that $p = \langle w, \Theta \rangle$ but $\Theta$ is adjusted to follow the order of $\xi$. 

\noindent \underline{\textbf{\textsc{Rule3}}}: If $w$ is involved in a proposition $\psi = \langle w, \xi\rangle$ where $\psi \in \Psi$, $\xi = \langle \tau(\theta'_0),\dots,\tau(\theta'_n)\rangle$, and there is a $\theta'$ lacking an one-to-one correspondence letting $ \tau(\theta')  = \tau(\theta)$, we consider $\xi$ is not an applicable topical proposition for $f$. We replace $w$ with a word with slightly lower probability until we can find an applicable topical proposition.

Following \underline{\textbf{\textsc{Rules}}}, we map each sentence in $t$ to propositions, and regard them as a state. All textual observations $\mathcal{T}$ are mapped to a state trace, denoted by $S$.

\vspace{-1.5mm}

\subsection{Action Models learning}


In this section, we address an action learner based on noisy state traces. We first use proposition extractor to generate state traces $S$ from textual observations $\mathcal{T}$ for learning. Likewise, goals $g$ are computed from $t_g$. Next, we use these extractions and plans to learn action models with the help of {\ARMS} \cite{ARMS}. However, the traces may be noisy, generated whether by the rough extractor mapping topics to propositions or the neural proposition extractor. To handle it, we learn various versions of action models based on each plans. Then we build a neural model inspired by matrix decomposition to get the most available action model. 


Specifically, we use {\ARMS} to learn $N$ versions of planning domains, denoted by  $\bar{\mathcal{D}}^n = \ARMS(S_n,\sigma_n,g_n)$. The reason for using each plan to learn is that numbers of noisy traces disturb learning. After attaining $\langle \bar{\mathcal{D}}^0, \dots, \bar{\mathcal{D}}^{N-1} \rangle$, we learn a conclusive planning domain by the action learner, denoted by $\dot{\mathcal{D}}$ = $\ActionLearner(\bar{\mathcal{D}}^0, \dots, \bar{\mathcal{D}}^{N-1}, \Psi)$. For each $A_i\in \mathcal{D}$ requiring complements, it relates to $K$ learned action models, i.e., $\langle \hat{A}_i^0, \dots, \hat{A}_i^{N-1} \rangle$. According to $\Psi$, recording extracted topical propositions, we use two $N \times L$ matrices, $M^p_{i}$ and $M^e_{i}$, to encode learned preconditions and effects respectively. $L$ is the quantity of extracted topical propositions.

To construct precondition matrix $M_i^p$, each row is an one-hot vector for $\hat{A}_i^{0:N-1}$ to indicate whether a topical proposition is included in their $\pre(a)$ or not. Similarly, $M_i^e$ indicates $\eff(a)$. Ideally, propositions are extracted correctly, letting action models learned from different state traces be the same. In this way, every row of $M_i^p$ is identical. $M_i^p$ can be decomposed as a product of two vectors, denoted by $M_i^p = \alpha\beta^\top$ where $\alpha$ is an $\textbf{1}_N$ vector and $\beta$ is a $N$-dimensional binary vector. Moreover, $\beta$ equals any row of $M_i^p$, i.e., $\pre(a)$. Similarly, we decompose $M_i^e$ by $M_i^e = \alpha\eta^\top$, where $\eta$ stands for $\eff(a)$. The objective is to compute $\beta$ and $\eta$ for a minimal loss, where $\beta$ and $\eta$ stand for the most available preconditions and effects. The loss is computed by Equation (\ref{equation:loss_action_model}). 
\vspace{-3mm}
\begin{align}
    L_a = \sum_{i=0}^{x}(||M^{p}_i - \alpha \beta_i^\top||_F + ||M^{e}_i - \alpha \eta_i^\top||_F) 
    \label{equation:loss_action_model}
\end{align}
\vspace{-4mm}\\
where $\langle \beta_0,\dots, \beta_x \rangle$ and $\langle \eta_0,\dots,\eta_x \rangle$ indicate preconditions and effects for each action model. For each topical proposition $\psi^j \in \Psi$, if $\beta_i^j > \lambda$, we consider that $\psi^j$ is a precondition of action model $A_i$, where $\lambda$ a predifined positive threshold. Similar, we complete effects of $A_i$ by $\eta_i$. In this way, we construct a conclusive planning domain $\dot{\mathcal{D}}$.

\vspace{-1mm}

\subsection{Neural Proposition Extractor}
Gained conclusive action models $\dot{\mathcal{D}}$
, we can train a neural proposition extractor by handling the contradiction between extracted traces and updated traces according to $\dot{\mathcal{D}}$.

\noindent \textbf{Training Data Building}
Given learned action models, we compare extracted state traces to state traces generated following $\dot{\mathcal{D}}$ to train proposition extractor to avoid conflicts. We denote the dataset by $\Pi$, each $\pi = \langle f, \rho, \Theta \rangle$. 
We first use $\Pi$ to collect all sentences and their extracted propositions. Next, as for each plan $\sigma$ and an initial state $s_0$ extracted from $t_0$, we let $\dot{s}_0 = s_0$ and compute each state $\dot{s}_{i+1}$ by executing actions $a_i$ in $\sigma$ to update $\dot{s}_i$, each action is based on $\dot{\mathcal{D}}$. We define an updated state trace by $\dot{S} = \langle \dot{s}_0, \dot{s}_1, \dots, \dot{s}_n\rangle$. 
Starting from state $s_1$, extracted by {\our}, we first compare it to $\dot{s}_1$. As for a sentence $f \in t_1$, its extracted proposition is $p = \langle \rho,\Theta \rangle $ where $p \in s_1$. If an unique proposition $\dot{p} = \langle \dot{\rho},\dot{\Theta} \rangle$ where $\dot{p} \in \dot{s}_1$, satisfies that $\dot{\rho} = \rho$ and there is an one-to-one correspondence letting all $\tau(\dot{\theta})  = \tau(\theta)$ but with different order, we substitute $\pi = \langle f, \rho, \dot{\Theta} \rangle$ into $\Pi$. On the other hand, if there is a one-to-one corresponding proposition $\dot{p} = \langle \dot{\rho},\dot{\Theta} \rangle$, where $\dot{p} \in \dot{s}_1$, satisfying $\dot{\Theta} = \Theta$ and $\dot{\rho} \neq \rho$, we substitute $\pi = \langle f, \dot{\rho}, \Theta \rangle$ into $\Pi$. After going through all traces, we complete a revised dataset for training neural proposition extractor.

\noindent \textbf{Neural Proposition Extractor}
After updating $\Pi$, we use it to train neural proposition extractor $\Extractor$, which is composed of a predicate generator $\Extractor_\rho$ and a parameter sequencer $\Extractor_\theta$. Predicate generator, composed of fully connected layers, first takes $h$ as an input, and outputs a probability distribution of each word to be predicate. According to Equation (\ref{equation:candidate_predicate}) for initializing propositions, we recompute the likelihood $P(w)$ of each word $w$ to be the predicate by Equation (\ref{equation:candidate_predicate_2}):
\vspace{-2mm}
\begin{align}
    \notag
    P(w) &= softmax(\Vocabulary \Topic^\top \Extractor_\rho(h)) \\ &= softmax(\Vocabulary \Topic^\top \Extractor_\rho(\Encoder(f)))
    \label{equation:candidate_predicate_2}
\end{align}
\vspace{-5.5mm}

Parameter sequencer $\Extractor_\theta$ is a GRU layer with a fully connected layer, taking $h$ and $\Theta$ as inputs. The output of sequencer is the index of the parameter $\theta$ appearing, composed of one-hot vectors, whose dimension is the maximal number of parameters of all actions. We denote it by $\hat{I} = \Extractor_\theta(h,\Theta)$.

The loss function is composed of two parts, the first one is to minimize a reconstruction loss $|\rho - \hat{\rho}|$ in terms of cross-entropy, to let generated predicate $\hat{\rho}$ and label predicate $\rho$ be as identical as possible. Similarly, the other one is to minimize $|I - \hat{I}|$, where $I$ indicates the order of parameters appearing. Especially, when training extractor, we also update encoder to get a more available representation of sentence.

During using $\Extractor$, we first employ the predicate generator to compute $P(w)$ and parameter sequencer to compute $\hat{I}$, respectively. Next, we arrange $\Theta$ according to its predicted order $\hat{I}$. Finally, we reinitialize an empty topical propositions set $\Psi$ and extract proposition by Equation (\ref{equation:rules}).

\vspace{-1.5mm}
\subsection{ Overview of {\our}}
\vspace{-3mm}
\begin{algorithm}[!ht]
\caption{An overview of {\our}}
\label{algorithm:code}
\textbf{input:} $\Phi = \langle \phi_0,\dots,\phi_{N-1} \rangle$, $\mathcal{D}$\\
\textbf{output: $\dot{\mathcal{D}}$} 
\begin{algorithmic}[1]
\STATE Train VAE by minimizing Equation (\ref{equation:loss});
\STATE Compute $\langle S_0, \dots, S_{N-1} \rangle$, $\langle g_0, \dots, g_{N-1} \rangle$ and $\Psi$;
\STATE Learn action models $\dot{\mathcal{D}} = \ActionLearner(\bar{\mathcal{D}}^0, \dots, \bar{\mathcal{D}}^{N-1},\Psi)$;
\FOR{iteration = 1, 2, ...}
\STATE Construct $\Pi$;
\STATE Train neural proposition extractor $\Extractor$ with $\Pi$;
\STATE Compute $\langle S_0, \dots, S_{N-1} \rangle$, $\langle g_0, \dots, g_{N-1} \rangle$ and $\Psi$;
\STATE Learn action models $\dot{\mathcal{D}} = \ActionLearner(\bar{\mathcal{D}}^0, \dots, \bar{\mathcal{D}}^{N-1},\Psi)$;
\ENDFOR
\end{algorithmic}
\end{algorithm}
\vspace{-4mm}
An overview of {\our} is shown in Algorithm \ref{algorithm:code}. We first train a VAE combined with a topic model (Line 1). Then we use it to encode sentences and initialize state traces and goals. During extracting, $\Psi$ records captured topical propositions (Line 2). 
Next, given initialized traces, goals and plans, we learn a planning domain $\dot{\mathcal{D}}$ with action learner $\ActionLearner$ (Line 3). 
After that, we repeat constructing $\Pi$ for training proposition extractor and learning action models until we get an optimal planning domain (Lines 4 to 9).

\vspace{-2mm}
\section{Experiments}
\subsection{Dataset}
We evaluate our approach {\our} on three domains: \emph{blocks}, \emph{minecraft}, and \emph{baking}. 
(1) \textbf{Blocks} The blocks domain describes how to pick and place five blocks on a table by a robot hand. (2) \textbf{Minecraft} The minecraft domain is about operating resources by an agent, such as crafting planks, picking up logs, and storing them in repositories. 
(3) \textbf{Baking} In the baking domain, an agent aims at baking cakes and souffles with eggs and a bag of flour. 
For each domain, we first constructed a planning domain model represented by PDDL. We then randomly generated 11,000 planning problems, with 10,000 for training and 1,000 for testing. We used planner \emph{Fast-downward} \cite{fast-downward} to generate propositional state traces by solving the 11,000 planning problems. After that, for each topical proposition, we constructed five templates for generating diverse sentences. We transformed each proposition to a sentence based on a template randomly selected from the five templates. 

\ignore{
\begin{itemize}
    \item \textbf{Blocks} The blocks domain describes how to pick and place five blocks on a table by a robot hand. 
    \item \textbf{Minecraft} The minecraft domain is about operating resources by an agent, such as crafting planks, picking up logs, and storing them in repositories. 
    \item \textbf{Baking} In the baking domain, an agent aims at baking cakes and souffles with eggs and a bag of flour. 
\end{itemize}
}

\vspace{-3mm}
\begin{table}[!ht]
\centering
\begin{tabular}{rrrr}
\toprule
                     & Blocks & Minecraft & Baking \\
                     \hline
\rule{0pt}{3.5mm} \scriptsize \makecell[r]{Topical  propositions } & 5      & 8         & 14     \\ 
\rule{0pt}{3.5mm} \scriptsize \makecell[r]{Action  models}        & 4      & 5         & 8      \\ 
\rule{0pt}{3.5mm} \scriptsize \makecell[r]{Types of  parameters}  & 5      & 8         & 14     \\ 
\rule{0pt}{3.5mm}\scriptsize \makecell[r]{Number of  parameters} & 6      & 22        & 19     \\ 
\rule{0pt}{3.5mm} \scriptsize Propositions          & 36     & 157       & 95    \\
\rule{0pt}{3.5mm} \scriptsize Words in vocabulary          &  41      &    71    &  105    \\
\rule{0pt}{3.5mm} \scriptsize Number of Sentences          &  553817      & 902632       &  1291727    \\
\bottomrule
\end{tabular}
\vspace{-1.5mm}
\caption{Features of three domains.}
\label{table:domain}
\vspace{-3mm}
\end{table}

We describe the following features of the three domains in Table \ref{table:domain}: (1) the number of topical propositions, (2) the number of action models, (3) the number of types of parameters, (4) the number of parameters in domain, (5) the number of propositions, (6) the size of vocabulary, 
(7) the number of sentences. Notably, we have deleted parameters from vocabulary. The three domains based on different scales focus on the size of propositions or action models.

In particular, most templates in the blocks domain have different ways of writing. Differently, the minecraft domains contains lots of similar sentences, such as ``You find a grass Grass0'' and ``You see a log Log1''. Compared to these domains, repeated expressions in the baking domains are more than blocks's but less than minecraft's. However, it contains the most quantity of topical propositions.

\vspace{-4mm}
\begin{table*}[!ht]
\centering
\setlength{\tabcolsep}{5mm}
\vspace{-3mm}
\begin{tabular}{rrrrr|rrrr}
\toprule
          & \multicolumn{4}{c}{$\ER$} & \multicolumn{4}{c}{$\RR$} \\
          & {\our}    & {\SPM}    & {\TMM}    & {\ablation}  & {\our}    & {\SPM}    & {\TMM}    & {\ablation}\\
\hline
Block     & \textbf{0.28} & 0.70 & 0.29 & 0.92 & 0.42 & \textbf{0.33} & 0.50 & 0.62 \\
Minecraft &\textbf{ 0.23} & 0.79 & 0.78 & 0.92 & \textbf{0.18} & 0.23 & 0.23 & 0.62 \\
Baking    & \textbf{0.16} & 0.77 & 0.35 & 0.98 & \textbf{0.49} & 0.63 & 0.68 &\textbf{0.49}   \\
\bottomrule
\end{tabular}
\caption{The error rate and redundancy rate of learned action models.}
\label{table:rate}
\vspace{-4mm}
\end{table*}

\vspace{1mm}
\subsection{Experimental Details}
In this paper, we trained (1) VAE for 3 epochs with learning rate $10^{-2}$ and batch size 32 (sentences), (2) action learner for 25000 epochs with learning rate $10^{-2}$, (3) neural proposition extractor for 3 epochs with learning rate $10^{-4}$ and batch size 32. And we repeat improving extractor and learning action models for 100 iterations. We ran all of our experiments on a machine with Ubuntu 16.04 on a 256GB of memory and a 12GB of GeForce GTX 1080 Ti. For space limitation, we show the network details in the supplementary material. 

We evaluate {\our} by comparing against three methods:

\noindent \textbf{Dependency Syntactic Parsing Method} 
Previous researches \cite{DBLP:conf/aips/LindsayRFHPG17,DBLP:conf/aaai/HaytonPFL20} about learning action models from texts describing actions, mostly first used Standford CoreNLP \cite{CoreNLP} to parse sentences, and then constructed action models based on some rules. Similarly, we built a dependency syntactic parsing method, {\SPM} for short, to handle textual observations.
Specifically, we first used CoreNLP dependency graph to capture syntactic parsing annotations, the root word of which is viewed as the predicate of a proposition. 
Parameters were extracted according to the order in sentences. After that, we constructed propositions based on Equation (\ref{equation:rules}) and used {\ARMS} to learn action models.

\noindent \textbf{Topic Model Method} 
{\TMM} (i.e., Topic Model Method) is the same as {\our} but without the EM-style framework. 

\noindent \textbf{{\our} w/o Guidance from Initializer}
We evaluated the effect of initializer by {\ablation}, which is the same as {\our} except for extracting with untrained proposition extractor $\Extractor$ directly instead of using initializer. Notably, at the beginning of learning action models, extracted state are almost random so that learner is hard to find a conclusive available planning domain. Once the learner cannot generate a valid domain, {\ablation} stops and outputs the final action models.




We evaluate {\our} with respect to the following aspects: 

(1) \textbf{Error Rate} The error rate $\ER$ is defined by the proportion of preconditions that cannot be established by any action in the previous part of the plan. (2) \textbf{Redundancy Rate} The redundancy rate $\RR$ is defined by the proportion of predicates in actions’ add list that does not establish any preconditions of any action in later part of the plan. (3) \textbf{Rand Index} Rand index $\RI$ is a measure to compare two data clusterings, for evaluating performance about classifying sentences into topical propositions.
\ignore{
\begin{itemize}
    \item \textbf{Error Rate} The error rate $\ER$ is defined by the proportion of preconditions that cannot be established by any action in the previous part of the plan.
    \item \textbf{Redundancy Rate} The redundancy rate $\RR$ is defined by the proportion of predicates in actions’ add list that does not establish any preconditions of any action in later part of the plan.
    \item \textbf{Rand Index} Rand index $\RI$ is a measure to compare two data clusterings, for evaluating performance about classifying sentences into topical propositions.
\end{itemize}}
\vspace{-3mm}

\subsection{Experimental Results}
\begin{figure*}[!b]
\vspace{-5mm}
    \centering
    \includegraphics[width=0.8\textwidth]{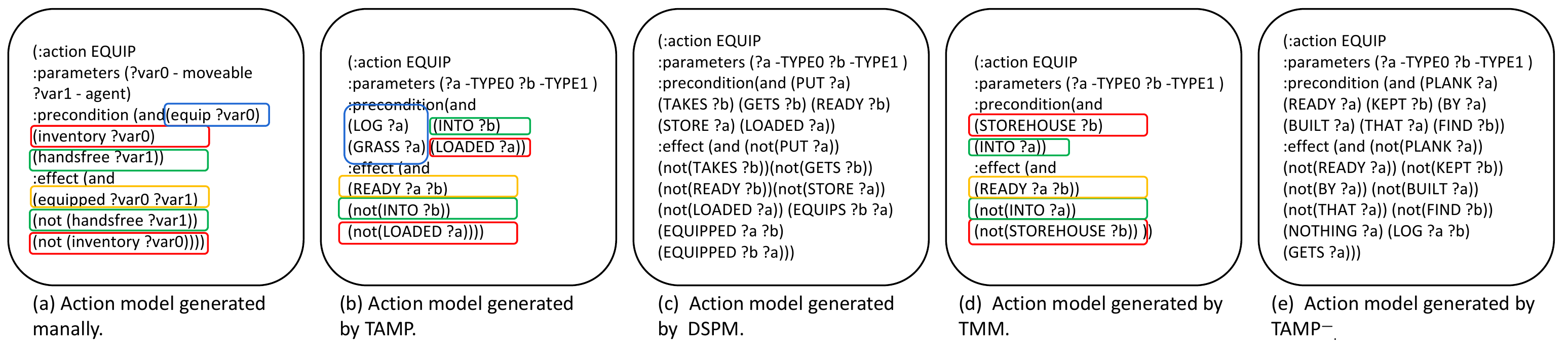}
    \vspace{-2.5mm}
    \caption{An example action model ``Equip (?var0 - moveable ?var1 - agent)'' in the minecraft domain.}
    \label{fig:case_action_model}
\end{figure*}

\noindent \textbf{Error Rate and Redundancy Rate}
To investigate the quality of learned action models, we tested the error rate $\ER$ and redundancy rate $\RR$. As shown in Table \ref{table:rate}, {\our} shows the best performance in all three domains with respect to $\ER$, indicating {\our} can indeed capture more accurate regularities compared to other approaches. In particular, the reason for the lower performance of {\SPM} is that it relies on whether sentences involve the same predicates or not. {\TMM} performs well in the blocks and baking domains, because it is able to assign sentences to topics instead of similar ones. {\our} makes up for it by using EM-style framework to iteratively train a neural extractor and learn action models for the optimal action models. {\ablation} has the worst performance, because, given random state traces, it is too hard for {\ablation} to learn action models, indicating the effectiveness of the initializer in {\our}.

Redundancy rate $\RR$ not only evaluate action models, but also exhibit the quality of extraction. The more the classified propositions are, the larger the $\RR$ is. As shown in Table \ref{table:rate}, {\our} obtains the best performance in minecraft and baking domains. The reason why {\our} performs slightly worse in the blocks domain is because sentences with the same meaning are represented in many different ways. 
{\our} fails gathering them into groups and includes all of them into action models to guarantee validity. {\SPM} directly uses words involved in the sentences to form propositions, which is difficult to estimate whether the word stands for the main idea or not.
Especially, the minecraft domain contains many similar sentences written in alike ways. Therefore, $\RR$ in minecraft domain is less than the other domains.



\noindent \textbf{Rand Index} 
We also evaluated the performance of classifying sentences into topical propositions by Rand Index $\RI$, as shown in Table \ref{table:rand}. The results exhibit that {\our} is superior to other approaches, indicating its ability to classify sentences into more accurate topical propositions. Compared to {\TMM}, the improvements indicate that the EM-style framework can indeed improve the performance of both extractor and action models learning.

\vspace{-3mm}
\begin{table}[!h]
\centering
\setlength{\tabcolsep}{3.5mm}
\begin{tabular}{rrrrr}
\toprule
          & {\our}    & {\SPM}    & {\TMM}    & {\ablation}  \\
\hline
Block     & \textbf{0.91} & 0.86 & 0.87 & 0.70 \\
Minecraft & \textbf{0.86} & 0.80 & 0.82 & 0.77 \\
Baking    & \textbf{0.94} & 0.70 & 0.92 & 0.89    \\
\bottomrule
\end{tabular}
\vspace{-1mm}
\caption{Rand index of extracted propositions.}
\label{table:rand}
\vspace{-3mm}
\end{table}

\noindent \textbf{Case Study} Figure \ref{fig:case_action_model} shows an example action model in the minecraft domain. Despite that {\our} does not gather the two propositions circled by blue blocks into one, its learned action model is the most complete compared with action models generated by the other approaches.

\vspace{-3mm}
\section{Conclusion}
In this paper, we present a novel approach {\our} to capture information from textual observations and learn action models from noisy traces. The EM-style framework enables {\our} to achieve a balance by iteratively improving extractor and action models. By conducting experiments on three domains, the experimental results show the superiority of {\our}. To the best of our knowledge, this is the first work on learning action models from textual observations. In the future, we will extend our approach into handling more realistic and complex descriptions and incomplete or noisy observations. 
\bibliographystyle{named}
\bibliography{ijcai21}

\end{document}